\documentclass[acmtog]{acmart}
\acmSubmissionID{442}

\usepackage{booktabs} 

\usepackage[ruled]{algorithm2e} 

\SetAlFnt{\small}
\SetAlCapFnt{\small}
\SetAlCapNameFnt{\small}
\SetAlCapHSkip{0pt}

\acmJournal{TOG}

\begin{teaserfigure}
  \includegraphics[width=1\textwidth]{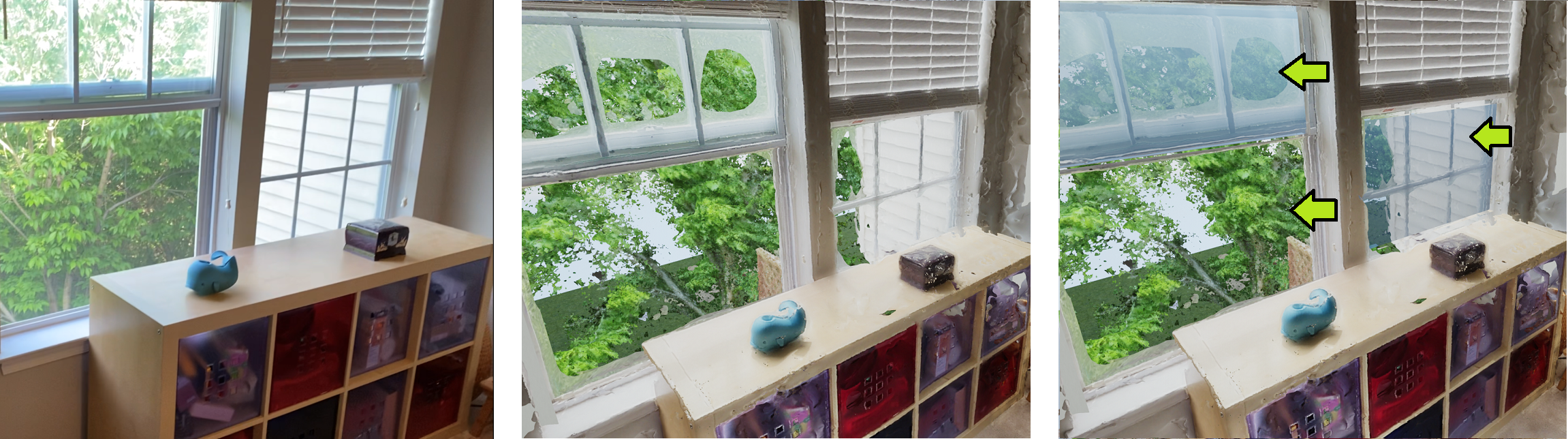}
  \vspace*{-2em}
  \caption{~\textit{Left}: ground truth image. Before (\textit{Middle}) and after (\textit{Right}) audio-augmented rendering of an indoor scene with an open and closed reflective surfaces. The reconstruction is enhanced by EchoCNN inferences of surface detection, depth estimation, and material classification based on audio-visual reflecting sound and image inputs. Green arrows highlight areas enhanced by our method. Glass surfaces added for closed windows and open areas remain unfilled.
  }
  \label{fig:one}
  \vspace*{0.5em}
\end{teaserfigure}

\begin{document}
\title{Echo-Reconstruction: Audio-Augmented 3D Scene Reconstruction 
}

\author{Justin Wilson}
\affiliation{%
  \institution{University of North Carolina at Chapel Hill}
  \city{Chapel Hill}
  \state{NC}
  \postcode{27599}
  \country{USA}}
\email{wilson@cs.unc.edu}
\author{Nicholas Rewkowski}
\affiliation{%
  \institution{University of Maryland at College Park}
  \city{College Park}
  \state{MD}
  \postcode{20740}
  \country{USA}}
\email{nick1@umd.edu}
\author{Ming C. Lin}
\affiliation{%
 \institution{University of Maryland at College Park}
  \city{College Park}
  \state{MD}
  \postcode{20740}
  \country{USA}}
\email{lin@cs.umd.edu}
\author{Henry Fuchs}
\affiliation{%
  \institution{University of North Carolina at Chapel Hill}
  \city{Chapel Hill}
  \state{NC}
  \postcode{27599}
  \country{USA}}
\email{fuchs@cs.unc.edu}

\renewcommand\shortauthors{Wilson, J. et al}

\begin{abstract}
Reflective and textureless surfaces such as windows, mirrors, and walls can be a challenge for object and scene reconstruction. These surfaces are often poorly reconstructed and filled with depth discontinuities and holes, making it difficult to cohesively reconstruct scenes that contain these planar discontinuities. We propose ''{\em Echoreconstruction}'', an audio-visual method that uses the reflections of sound to aid in geometry and audio reconstruction for virtual conferencing, teleimmersion, and other AR/VR experience. The mobile phone prototype emits pulsed audio, while recording video for RGB-based 3D reconstruction and audio-visual classification. Reflected sound and images from the video are input into our audio (EchoCNN-A) and audio-visual (EchoCNN-AV) convolutional neural networks for surface and sound source detection, depth estimation, and material classification. The inferences from these classifications enhance scene 3D reconstructions containing open spaces and reflective surfaces by depth filtering, inpainting, and placement of unmixed sound sources in the scene. Our prototype, VR demo, and experimental results from real-world and virtual scenes with challenging surfaces and sound indicate high success rates on classification of material, depth estimation, and closed/open surfaces, leading to considerable visual and audio improvement in 3D scenes (see~\autoref{fig:one}).
\end{abstract}

%
%


%
%


\maketitle

\section{Introduction}
\label{Introduction}

Scenes containing open and reflective surfaces, such as windows and mirrors, can enhance AR/VR immersion in terms of both graphics and sound; for example, a window open in spring compared to closed in winter. However, they also present a unique set of challenges. First, they are difficult to detect and reconstruct due to their transparency and high reflectivity. Distinguishing between glass (e.g. window) and an opening in the space is an important part of the audio-visual experience for AR/VR engagement. Also, illumination, background objects, and min/max depth ranges can be confounding factors. 

Reconstruction of scenes for teleimmersion have led to advances in detection~\cite{DBLP:journals/corr/LeaFVRH16}, segmentation~\cite{Golodetz2015,Arnab_2015}, and semantic understanding~\cite{song2016ssc} and are used to generate large-scale, labeled datasets of object~\cite{Wu2015} and scene ~\cite{dai2017scannet} geometric models to further aid training and sensing in a 3D environment. Advances have also been made to account for challenging surfaces
~\cite{Sinha2012ImagebasedRF,Whelan:2018:RSM:3197517.3201319,DBLP:journals/corr/abs-1904-02251}. Yet, scenes containing open and reflective surfaces, such as windows and mirrors, remain an open research area. Our work augments existing visual methods by adding audio context of surface detection, depth, and material estimation for recreating a virtual environment from a real one.

Previous work has used sound to better understand objects in scenes. For instance, impact sounds from interacting with objects in a scene to perform segmentation~\cite{Arnab_2015} and to emulate the sensory interactions of human information processing~\cite{gensound}. Audio has also been used to compute material~\cite{Ren:2013:EPB:2421636.2421637}, object~\cite{gensound}, scene~\cite{Schissler2018AcousticCA}, and acoustical~\cite{tang2019sceneaware} properties. Moreover, using both audio and visual sensory inputs has proven more effective; for example, multi-modal learning for object classification~\cite{10.1007/978-3-030-01267-0_34,Wilson19} and object tracking~\cite{Wilson20_icra}.

Fusing multiple modalities, such as vision and sound, provide a wider range of possibilities than either single modality alone. In this work, we show that augmenting vision-based techniques with audio, referred to as ``EchoCNN,'' can detect open or reflective surfaces, its depth, and material, thereby enhancing 3D object and scene reconstruction for AR/VR systems. We highlight key results below:

\begin{itemize}
    \item EchoReconstruction, a staged audio-visual 3D reconstruction pipeline that uses mobile devices to enhance scene geometry containing windows, mirrors, and open surfaces with depth filtering and inpainting based on EchoCNN inferences (\autoref{TechnicalApproach});
    \item EchoCNN, a fused audio-visual CNN architecture for classifying open/closed surfaces, their depth, and material or sound source placement (\autoref{ModelArchitecture});
    \item Automated data collection process and audio-visual ground truth data for real and synthetic scenes containing windows and mirrors (\autoref{Datasets}).
\end{itemize}

Using EchoReconstruction, we have been able to achieve consistently higher accuracy (up to 100\%) in classification of open/closed surfaces, depth estimation, and materials in both real-world scenes and controlled experiments, resulting in considerably improved 3D scene reconstruction with glass doors, windows and mirrors.

\begin{figure}
  \includegraphics[width=0.5\textwidth]{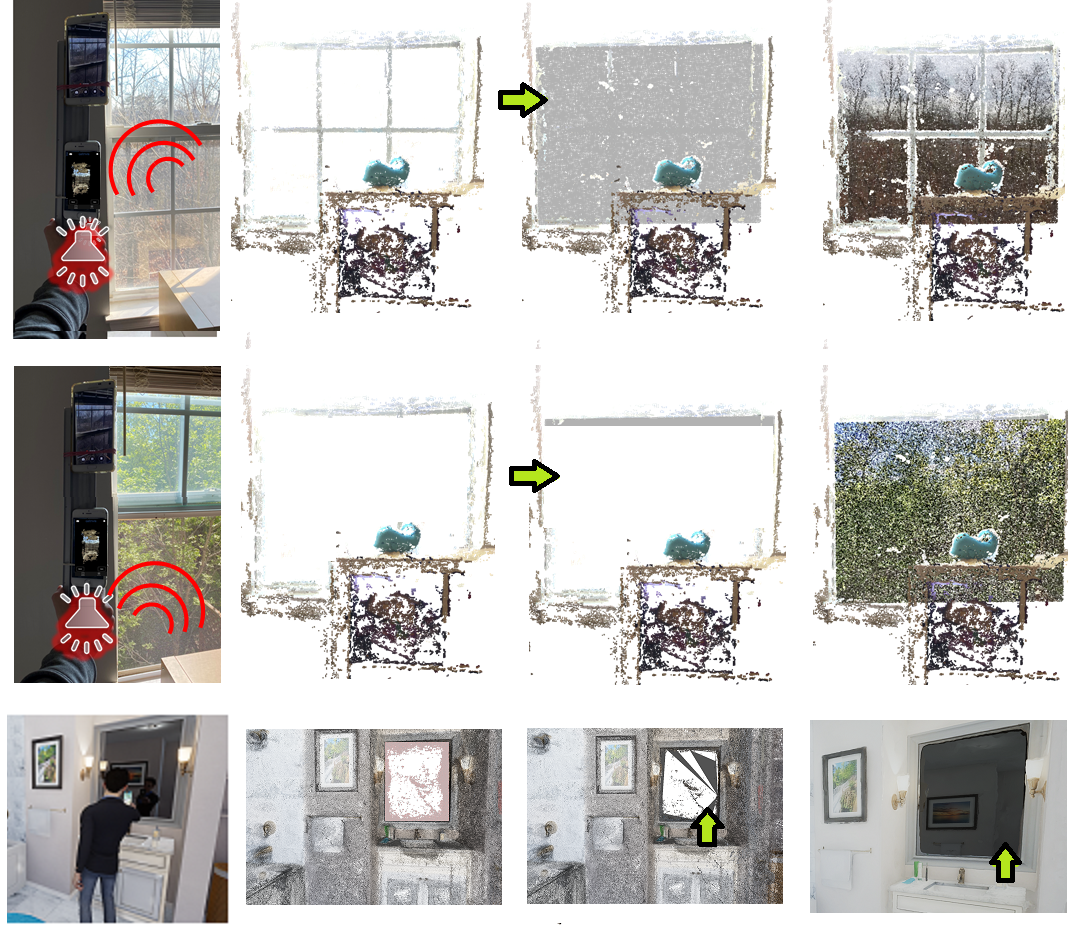}
  \vspace*{-2em}
  \caption{\textit{Top row}: closed window in winter.~\textit{Middle row}: opened in spring.~\textit{Bottom row}: controlled experiment virtual scene.~\textit{Column 1}: mobile echoreconstruction prototype in real world and virtual scenes with a video of emitted pulsed audio and images from the scene.~\textit{Column 2}: initial RGB based 3D reconstruction using state-of-the-art visual methods (live~\cite{Tanskanen2013LiveM3} or photogrammetric~\cite{Metashape20}).~\textit{Column 3}: our audio-visual EchoCNN convolutional neural network classifies open or closed surface, depth, and material for inpainting to resolve planar discontinuities caused by reflective surfaces, such as windows and mirrors.~\textit{Column 4}: semantic rendering of the window given material estimation and point of view. Green arrows highlight areas enhanced by our method, such as detecting closed/open parts of a window and filling a reflective mirror.
  }
  \label{fig:pointCloud}
  \vspace*{-1em}
\end{figure}

\section{Related Work}
\label{RelatedWork}
Previous research in 3D reconstruction, audio-based classifications, and echolocation are discussed in this section in addition to existing techniques for reconstructing open and reflective surfaces.

\subsection{3D reconstruction}
\label{3DReconstruction}

Object and scene reconstruction methods generate 3D scans using RGB and RGB-D data. For example, Structure from Motion (SFM)~\cite{WESTOBY2012300}, Multi-View Stereo (MVS)~\cite{Seitz2006}, and Shape from Shading~\cite{784284} are all techniques to scan a scene and its objects. Static~\cite{Newcombe2011,Golodetz2015} and dynamic~\cite{Newcombe2015,Dai2017} scenes can also be scanned in real-time using commodity sensors such as the Microsoft Kinect and GPU hardware. 3D scene reconstructions have also been performed with sound based on time of flight sensing~\cite{DBLP:journals/corr/CroccoTB16}. Not only has this previous research generated large amounts of 3D scene~\cite{Silberman:ECCV12,song2016ssc} and object~\cite{Singh2014,Lai2011,Wu2015} data, they also benefit from these datasets by using them for training vision-based neural networks for classification, segmentation, and other downstream tasks. Depth estimation algorithms~\cite{DBLP:journals/corr/EigenF14,Alhashim2018,DBLP:journals/corr/abs-1904-02251} also create 3D reconstructions by fusing depth maps using ICP and volumetric fusion~\cite{10.1145/2047196.2047270}.

\subsubsection{Glass and mirror reconstruction}
\label{GlassAndMirrorReconstruction}
Reflective surfaces produce identifiable audio and visual artifacts that can be used to help their detection. For example, researchers have developed algorithms to detect reflections in images taken through glass using correlations of 8-by-8 pixel blocks~\cite{Shih:15}, image gradients~\cite{Kopf2013ImagebasedRI}, two layer renderings~\cite{Sinha2012ImagebasedRF}, polarization imaging reflectometry~\cite{Riviere2017PolarizationIR}, and diffraction effects~\cite{Toisoul2017}. Adding hardware, ~\cite{sutherland} used ultrasonic sensor logic to track continuous wave ultrasound and ~\cite{Zhang:17} to detect obstacles such as glass and mirrors by using frequencies outside of the human audible range. More recently, reflective surfaces have been detected by utilizing a mirrored variation of an AprilTag~\cite{Olson2011AprilTagAR,Wang2016AprilTag2E}.~\cite{Whelan:2018:RSM:3197517.3201319} use the reflective surface to their advantage by recognizing the AprilTag attached to their Kinect scanning device when it appears in the scene. Depth jumps and incomplete reconstructions have also been used~\cite{Lysenkov:08}. However, vision based approaches require the right illumination, non-blurred imagery, and limited clutter behind the surface that may limit the reflection. We show that sound creates a distinct audio signal, providing reconstruction methods complementary data about the presence of windows and mirrors without additional sensors.

\begin{table}
\centering
  \begin{tabular}{ll}
  \multicolumn{2}{c}{\bf Example 3D Reconstruction Methods} \\
  \hline
    Type & Methods \\ 
    \hline
    Active (RGB-D) & KinectFusion, DynamicFusion, BundleFusion \\
    Passive (RGB) & SLAM, SFM,~\cite{Tanskanen2013LiveM3} \\ 
     & ScanNet,~\cite{Whelan:2018:RSM:3197517.3201319} \\
    Stereo & MVS, StereoDRNet \\
    \hline
    Lidar & \cite{Kada20093DBR}   \\ 
    Ultrasonic & \cite{Zhang:17} \\ 
    Time of flight & \cite{DBLP:journals/corr/CroccoTB16} \\
  \end{tabular}
  \caption{3D reconstruction methods by type such as passive (RGB), active (RGB-D), or other sensor (e.g. ultrasound, lidar, etc.); single or multiple views; and static or dynamic scenes.
  }
  \vspace*{-2.5em}
  \label{tab:exampleReconstructionMethods}
\end{table}

\subsection{Acoustic imaging and audio-based classifiers}
\label{AcousticsAndClassification}
We begin with an introduction into sound propagation, room acoustics, and audio-visual classifiers.

\begin{figure*}
  \includegraphics[width=0.97\textwidth]{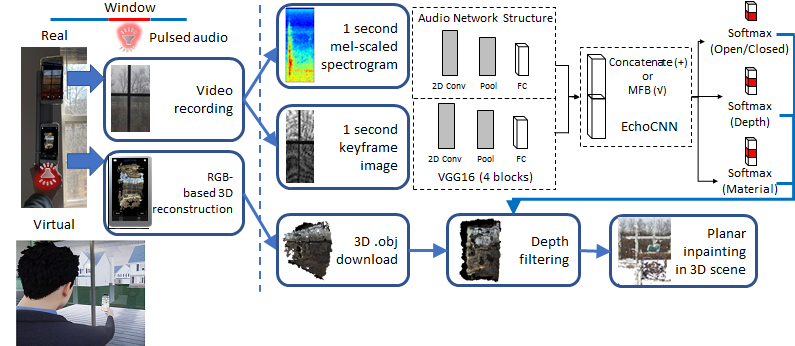}
  \vspace*{-1em}
  \caption{\textit{Staged approach} to enhance 3D virtual scene and object reconstruction using audio-visual data. Our echoreconstruction prototype consists of two smartphones - one recording (top) and one emitting/reconstructing (bottom). As the bottom smartphone moves to reconstruct the scene and emits 100 ms pulsed audio 
  , the top smartphone is used to record video of the direct and reflecting sound. The receiving audio is split into 1.0 second intervals to allow for reverberation. These audio intervals are converted into mel-scaled spectrograms and passed through a multimodal echoreconstruction convolutional neural network (we refer to as EchoCNN) comprised of 2D convolutional, max pooling, fully connected, and softmax layers. EchoCNN classifications inform depth filtering and hole filling steps to resolve planar discontinuities in scans caused by reflective surfaces, such as windows and mirrors. Binary classification is used to predict if a window is open or closed. Multi-class classification is used for depth and material estimation.
}
  \label{fig:stagedApproach}
  \vspace*{-0.5em}
\end{figure*}

\textbf{Acoustics}: various models have been developed to simulate sound propagation in a 3D environment, such as wave-based~\cite{Mehra:15}, ray tracing based~\cite{Rungta:2016:SIS:2913992.2914168}, sound source clustering~\cite{Tsingos:2004:PAR:1015706.1015710}, multipole equivalent source methods~\cite{James:2006:PAT:1179352.1141983}, and a single point multipole expansion method~\cite{Zheng:2011:THM:2010324.1964933}, representing outgoing pressure fields.~\cite{Godoy2018} uses acoustics and a smartphone for an app to detect car location and distance from walking pedestrians using temporal dynamics.~\cite{Bianco2019} further discusses theory and applications of machine learning in acoustics. Computational imaging approaches have also used acoustics for non-line-of-sight imaging~\cite{Lindell:2019:Acoustic}, 3D room geometry reconstruction from audio-visual sensors~\cite{Kim:17}, and acoustic imaging on a mobile device~\cite{10.1145/3210240.3210325}. To reconstruct windows and mirrors, our work uses room acoustics given the surface materials of the room~\cite{Schissler2018AcousticCA} and distance from sound source. However, prior work and downstream processes often require a watertight reconstruction which can be difficult to generate in the presence of glass.  Our approach addresses these issues using an integrated audio-visual CNN that can detect discontinuity, depth, and materials.

\textbf{Audio-based classification and reconstruction}: using principles from sound synthesis, propagation, and room acoustics, audio classifiers have been developed for environmental sound~\cite{45857,Piczak:2015:EDE:2733373.2806390,Salamon:2014:DTU:2647868.2655045}, material~\cite{Arnab_2015}, and object shape~\cite{gensound} classification. For audio-based reconstruction, Bat-G net uses ultrasonic echoes to train an auditory encoder and 3D decoder for 3D image reconstruction~\cite{Hwang2019BatGNB}. Audio input can take the form of raw audio, spectral shape descriptors~\cite{Buchler2005,COWLING20032895,josPhysicalAudioSignalProcessing}, or frequency spectral coefficients that we also adopt. In our method, we use reflecting sound to perform surface detection, depth estimation, and material classification.

\textbf{Audio-visual learning}: similar to its applications in natural language processing (NLP) and visual questing \& answering systems~\cite{kim2016multimodal,kim2020modalitybalanced,hannan2020manymodalqa}, multi-modal learning using both audio-visual sensory inputs has also been used for classification tasks~\cite{10.1007/978-3-030-01267-0_34,Wilson19}, audio-visual zooming~\cite{Nair:2019:AZY:3343031.3351010}, and sound source separation~\cite{DBLP:journals/corr/abs-1804-03619,Lee:2000:ANM:3008751.3008829} which have also isolated waves for specific generation tasks.  Although similar in spirit,
our audio-visual method, ``Echoreconstruction,'' differs from the existing methods by learning absorption and reflectance properties to detect a reflective surface, its depth, and material.

\section{Technical Approach}
\label{TechnicalApproach}
In this work, we adopt ``echolocation'' as an analog for our echoreconstruction method. According to~\cite{Egan1988}, echo is defined as \textit{distinct} reflections of the original sound with a sufficient sound level to be clearly heard above the general reverberation. Although perceptible echo is abated because of precedence (known as the Haas effect)~\cite{Long2014}, returning sound waves are received after reflecting off of a solid surface. We use these distinct, reflecting sounds to design a staged approach of audio and audio-visual convolutional neural networks. EchoCNN-A and EchoCNN-AV can be used to estimate depth based on reverberation times (\autoref{fig:activationMaximization}), recognize material based on frequency and amplitude, and handle both static and dynamic scenes with moving objects based on Doppler shift. All of which enhance scene and object reconstruction by detecting planar discontinuities from open or closed surfaces and then estimating depth and material.

\subsection{Echolocation}
\label{Echolocation}
Echolocation is the use of reflected sound to locate and identify objects, particularly used by animals like dolphins and bats. According to~\cite{SZABO2014501}, bats emit ultrasound pulses, ranging between 20-150 kHz, to catch an insect prey with a resolution of 2-15 mm. This involves signal processing such as:

\begin{enumerate}
    \item Doppler shift (the relative speed of the target),
    \begin{equation}
    \Delta f = f_D - f_0 = f_0\frac{c_s}{c_0}cos(\theta)
    \end{equation}
    \item time delay (distance to the target), and
    \item frequency and amplitude in relation to distance (target object size and type recognition);
\end{enumerate}

\noindent where the Doppler shift (or effect) is the perceived change in frequency (Doppler frequency $f_D$ minus transmitted frequency $f_0$) as a sound source with velocity $c_s$ moves toward or away from the listener/observer with velocity $c_o$ and angle $\theta$. 

\subsection{Staged classification and reconstruction pipeline}
\label{MultiStageApproach}
As depicted in~\autoref{fig:stagedApproach}, we take a staged approach to enhance scene and object reconstruction using audio-visual data. Our echoreconstruction prototype consists of two smartphones - one recording (top) and one emitting/reconstructing (bottom). Each audio emission is 100 ms of sound followed by 900 ms of silence to allow for the receiving microphone to capture reflections and reverberations (\autoref{SoundSource}). After the 3D scan is complete, an .obj file containing geometry and texture information is generated. 1 second frames are extracted from the recorded video to generate audio and visual input into the EchoCNN neural networks (\autoref{ModelArchitecture}). These networks are independently trained to detect whether a surface is open or closed, estimate depth to the surface from the sound source, and classify the material of the surface. Using mobile accelerometer data and a multi-scale neural network, such as~\cite{DBLP:journals/corr/EigenF14}, but using audio as a coarse global output refined using finer-scale visual data to augment depth estimation will be explored as future work.

\begin{figure}
  \includegraphics[width=0.47\textwidth]{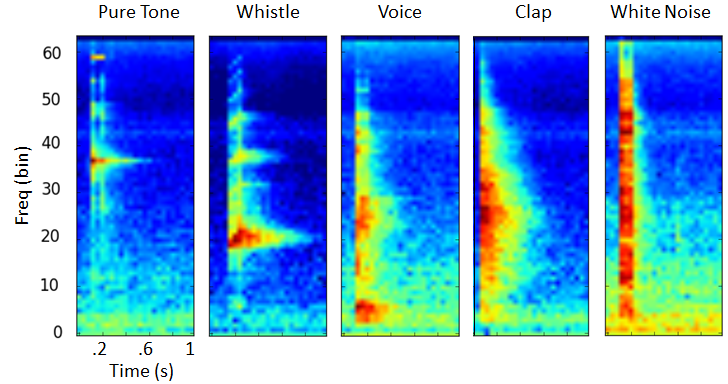}
  \vspace*{-1em}
  \caption{Mel-scaled spectrograms of recorded impulses of different sound sources used.~\textit{From left to right}: narrow to disperse spectra. Not shown are other pure tone frequencies, chirp, pink noise, and brownian noise. Horizontal axis is time and vertical axis is frequency.}
  \label{fig:soundSourceSpectrograms}
  \vspace*{-1em}
\end{figure}

\subsection{Sound source}
\label{SoundSource}
A smartphone emits recordings of human experimenter voice, whistle, hand clap, pure tones (ranging from 63 Hz to 16 kHz), chirps, and noise (white, pink, and brownian). All of which can be generated as either pulsed (PW) or continuous waves (CW). PW is preferred for theoretical and empirical reasons. First, the transmission frequency $f_0$ may experience considerable downshift as a result of absorption and diffraction effects~\cite{SZABO2014501}. Therefore, using pulsed waves independent for each emission is theoretically better than continuous waves compared to $f_0$. Furthermore, \autoref{Results} shows superior PW results over CW for the given classification tasks.

Pure tones were generated with default 0.8 out of 1 amplitudes using the Audacity computer program and center frequencies of 63 Hz, 125 Hz, 250 Hz, 500 Hz, 1 kHz, 2 kHz, 4 kHz, 8 kHz, and 16 kHz. Human voice ranges from about 63 Hz to 1 kHz~\cite{Long2014} (125 Hz to 8 kHz~\cite{Egan1988}) and an untrained whistler between 500 Hz to 5 kHz~\cite{Nilsson2008}. Chirps were linearly interpolated from 440 Hz to 1320 Hz in 100 ms. A hand clap is an impulsive sound that yields a flat spectrum~\cite{Long2014}. All sound sources were recorded and played back with max volume (\autoref{fig:soundSourceSpectrograms}). While recorded sounds were used for consistency, we plan to add live audio for augmentation and future ease of use during reconstruction. Please see our supplementary materials for spectrograms across all sound sources.

\textbf{Audio input}: audio was generated in pulsed waves (PW). One smartphone to emit the sound while performing a RGB-based reconstruction and the second smartphone to capture video. As future work, a single mobile device or Microsoft Kinect paired with audio chirps could be used for audio-visual capture and reconstruction instead of two separate devices. Each pulsed wave emitted into the scene was a total of 1 second consisting of an 100 ms impulse followed by silence. 1 second audio frames is based on the Sabine Formula of reverberation time for a compact room of like dimensions calculated as:
\begin{equation}
    T = 0.05\frac{V}{a} = 0.05\frac{V}{\sum S\alpha} = (0.05\frac{\textit{sec}}{\textit{ft}})\frac{1,296 \textit{ ft}^3}{69.23 \textit{ ft}^2} = 0.94 \textit{ sec}
\end{equation}
\label{eqn:reverbTime}
\noindent where $T$ is the reverberation time (time required for sound to decay 60 dB after source has stopped), $V$ is room volume ($\textit{ft}^3$), and $a$ is the total room absorption at a given frequency (e.g. 250 Hz). For the bathroom scene, $V = 9\textit{ ft} * 16\textit{ ft} * 9\textit{ ft} = 1,296\textit{ ft}^3$ and $a = 69.23\textit{ ft}^2$, which is the sum of sound absorption from the materials in~\autoref{tab:sabine}.

\begin{table}
\centering
  \caption{According to the Sabine Formula (\autoref{eqn:reverbTime}), reverberation time can be calculated as room volume V divided by total room absorption a. For an indoor sound source in a reverberant field, a is the total room absorption at a given frequency (sabins), S is the surface area ($\textit{ft}^2$), and $\alpha$ is the sound absorption coefficient at a given frequency (decimal percent). At 250 Hz, the total room absorption a for our real-world bathroom scene is 69.23 sabins.
  }
  \vspace*{-0.5em}
  \label{tab:sabine}
  \begin{tabular}{lccc}
  \multicolumn{4}{c}{Total room absorption a using $a = \sum S\alpha$ at 250 Hz} \\
  \hline
    Real bathroom scene & S & $\alpha$ & a (sabins) \\ 
    \hline
    Painted walls & 432 x & 0.10 = & 43.20  \\ 
    Tile floor & 175 x & 0.01 = & 1.75  \\
    Glass & 60 x & 0.25 = & 15.00  \\ 
    Ceramic & 39 x & 0.02 = & 0.78  \\ 
    Mirror & 34 x & 0.25 = & 8.50  \\ 
    \multicolumn{3}{r}{Total a =} & 69.23 sabins
  \end{tabular}
\end{table}

\textbf{Visual input}: images were captured from the same smartphone video as the audio recordings. Each corresponding image was cropped and grayscaled for illumination invariance and data augmentation. Image dimensions were 64 by 25 pixels. Visual data served as inputs for visual only and audio-visual model variation EchoCNN-AV.

\subsection{Initial 3D Reconstruction}
\label{Initial3DReconstruction}
We evaluated the following smartphone based reconstruction applications to obtain an initial 3D geometry for which our method would enhance. The Astrivis application, based on~\cite{Tanskanen2013LiveM3}, generates better live 3D geometries for closed object rather than scene reconstructions since it limits feature points per scan. On the other hand, Agisoft Metashape produces scene reconstructions offline from smartphone video. Enabling the software's depth point and guided camera matching features further improved reconstructed geometries.

\section{Model Architecture}
\label{ModelArchitecture}
To augment visually based approaches, we use a multimodal CNN with mel-scaled spectrogram and image inputs. First, we perform surface detection to determine if a space with depth jumps and holes is in error or in fact open (i.e. open/closed classification). In the event of error, we estimate distance from recorder to surface using audio-visual data for depth filtering and inpainting. Finally, we determine the material. All of these classifications are performed using our audio and audio-visual convolutional neural networks, referred to as EchoCNN-A and EchoCNN-AV (Fig.~\ref{fig:stagedApproach}).

\begin{figure}[th]
  \includegraphics[width=0.45\textwidth]{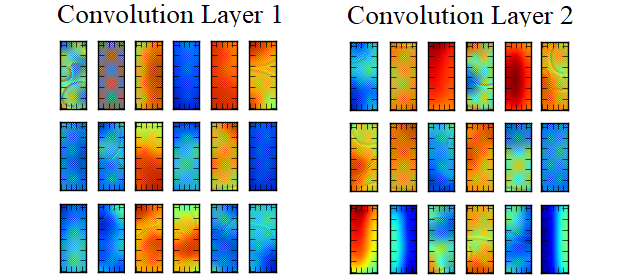}
  \vspace*{-0.5em}
  \caption{Sample visualizations of the filters for the two convolutional layers in the audio-based EchoCNN-A neural network. The model learns filters for octave bands, frequencies, reflections, reverberations, and damping.}
  \label{fig:convFilters}
  \vspace*{-1em}
\end{figure}

\textbf{Audio sub-network}: our frame-based EchoCNN-A consists of a single convolutional layer followed by two dense layers with feature normalization. Sampled at $F_s = 44.1\textit{ kHz}$ to cover the full audible range, audio frames are 1 second mel-scaled spectrograms with STFT coefficients $\chi$ (\autoref{eqn:stft}). Each audio example is classified independently and 1 second intervals to reflect an estimated reverberation time based on a compact room size (\autoref{eqn:reverbTime}). With a 2048 sample Hann window (N), 25\% overlap, and hop length ($H = 2048/4$), this results in a frequency dimension of 21.5 Hz (\autoref{eqn:freqDim}) and temporal dimension of 12 ms (\autoref{eqn:timeDim}) or 12\% of each 100 ms pulsed audio. Each spectrogram is individually normalized and downsampled to a size of 62 frequency bins by 25 time bins.

We define the frequency spectral coefficients \cite{M15} as:

\begin{equation}
    \chi(m,k) = \sum^{N-1}_{n=0} x(n+mH)w(n)exp(-2\pi ikn/N)
\end{equation}
\label{eqn:stft}

\noindent for $m^{th}$ time frame and $k^{th}$ Fourier coefficient with real-valued DT signal $x: \textit{Z} \rightarrow \textit{R}$, sampled window function $w(n) \textit{ for n} \in [0:N-1] \rightarrow \textit{R}$ of length $N \in \textit{N}$, and hop size $H \in \textit{N}$~\cite{M15}. $\textit{R}$ denotes continuous time and $\textit{Z}$ denotes discrete time. Equal to $|\chi(m,k)|^2$, spectrograms have been demonstrated to perform well as inputs into convolutional neural networks (CNNs)~\cite{DBLP:journals/corr/Huzaifah17}. Their horizontal axis is time and vertical axis is frequency.

\begin{equation}
    F_{coef}(k)=\frac{k\dot F_s}{N} = k\frac{44100}{2048} = k * 21.5 \textit{ Hz}
\end{equation}
\label{eqn:freqDim}
\begin{equation}
    T_{coef}(m) = \frac{m\dot H}{F_s} = m\frac{2048 * 0.25}{44100} = m * 0.012 \textit{ seconds}
\end{equation}
\label{eqn:timeDim}

A hop length of $H = N/2$ achieves a reasonable temporal resolution and data volume of generated spectral coefficients~\cite{M15}. Temporal resolution is important in order to detect when a reflecting sound reaches the receiver. Therefore, we decided to use a shorter window length $N=2048$ instead of $N=4096$ for instance. This resulted in a shorter hop length and accepting the trade-off of a higher temporal dimension for increased data volume.

\textbf{Visual sub-network}: while audio information is generally useful for all three classifications tasks (\autoref{tab:model-eval}) visual information is particularly useful to aid material classification. We use ImageNet~\cite{NIPS2012_4824} as a visual-based baseline to compare to our audio and audio-visual methods. It also serves as an input into our audio-visual merge layer. Future work will explore whether or not another image classification method is better suited as a baseline and to fuse with audio.

\begin{figure*}
  \includegraphics[width=0.99\textwidth]{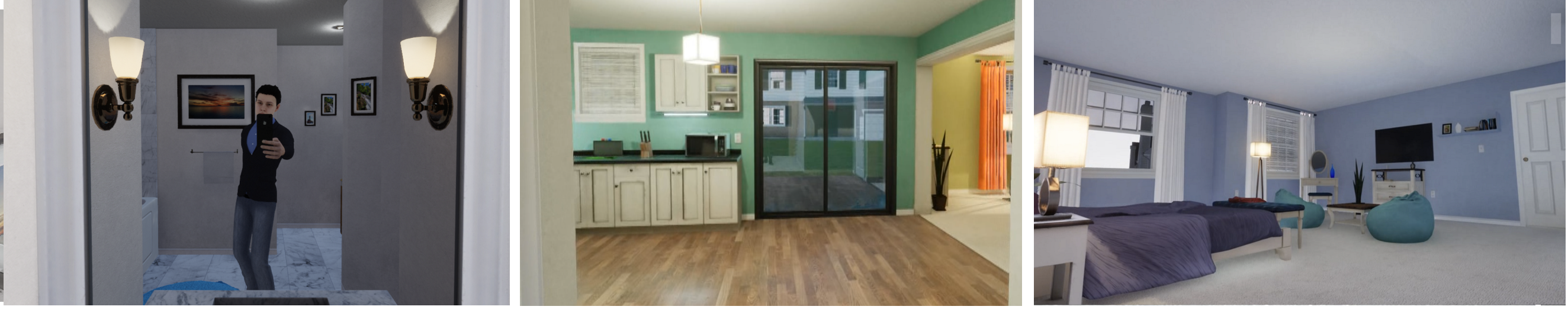}
  \vspace*{-1.5em}
  \caption{Listener at different distances from sound source (from 0.5 to 3 m) in a virtual environment (left: bathroom, middle: kitchen, right: bedroom) used to generate synthetic audio-visual data. This dataset is comprised of multiple 12-second video clips in front of reflective surfaces at increments from 0.5 m to 3 m for 15 different sound sources. Absorption and transmission coefficients were set on materials (e.g. mirror, thick glass, ordinary glass) inside and outside of rooms in the virtual scenes. In addition to open/closed, depth, and material, we make synthetic, unmixed reflection separation data (direct, early, or late) available for future research.
  }
  \label{fig:synDataset}
  \vspace*{-1em}
\end{figure*}

\begin{figure}
  \includegraphics[width=0.5\textwidth]{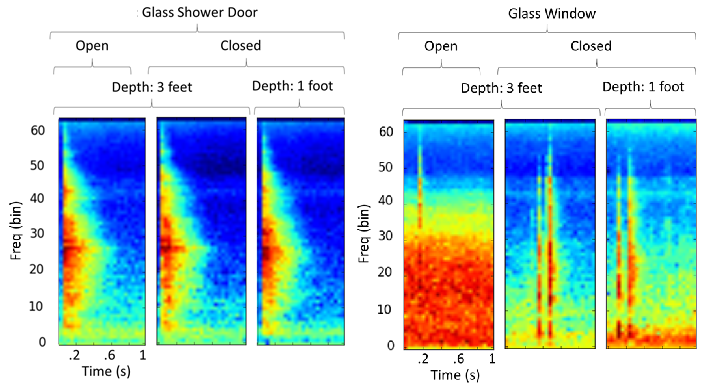}
  \vspace*{-2em}
  \caption{Spectrograms from a recorded hand clap in front of an interior glass shower door and exterior glass window. For the interior door, reflected sounds experience intensified damping as we go from opened (\textit{left}) to closed (\textit{middle}) and then from 3 feet to 1 foot depth (\textit{right}). Damping increases with fewer late reverberations and intensity increases with more early reflections. For the exterior window, closing it decreases outside noise up to a distance.}
  \label{fig:comparisonSpectrograms}
  \vspace*{-1.5em}
\end{figure}

\textbf{Merge layer}: we evaluated concatenation and multi-modal factorized bilinear (MFB) pooling~\cite{DBLP:journals/corr/abs-1708-01471} to fuse audio and visual fully connected layers. Concatenation of the two vectors serves as a straightforward baseline. MFB allows for additional learning in the form of a weighted projection matrix factorized into two low-rank matrices. 

\begin{equation}
    z_i = x^T W_i y
    = x^T U_i V_i^T y
    = \textit{1}^T (U_i^T x \textit{ } \circ \textit{ }V_i^Ty)
\end{equation}
\noindent where k is the factor or latent dimensionality with index i of the factorized matrices, $\circ$ is the Hadmard product or element-wise multiplication, and $\textit{1} \in \textit{R}^k$ is an all-one vector.

\subsection{Loss Function}
For open/closed predictions, categorical cross entropy loss (\autoref{eqn:crossEntropy}) is used instead of binary if estimating the extent of the surface opening (e.g. all the way open, halfway, or closed). A regression model is not used for depth estimation because ground truth data is collected in discrete 0.5 m or 1 ft increments within the free field for better noise reduction~\cite{Egan1988}. The Softmax function is used for output activations.

\begin{equation}
    -\sum_{c=1}^M y_{o,c} \log(p_{o,c})
\end{equation}
\label{eqn:crossEntropy}

\noindent where M is number of classes, y indicator for correct classification, and p for predicted probability that observation (o) is of class (c).

\begin{algorithm}[t]
\SetAlgoNoLine
\KwIn{EchoCNN classifications (open/closed, depth, and material) and initial 3D reconstruction (.obj and.jpg).}
\KwOut{Enhanced 3D echoreconstruction (updated .obj').}

\textbf{Variables}: 
.obj = initial 3D reconstruction, .obj' = enhanced 3D echoreconstruction, $i = classification_i$ based on EchoCNN inputs, $j = discontinuity_j$, $k = vertex_k$ in .obj'

\hspace{1cm}

.obj' = .obj

\For{each $classification_{i}$ of audiovisual input
}
{
    \If{(($classification_{i}[open/closed]$ == $closed$) \text{and} ($classification_{i}[material]$ in
    ($glass$,$mirror$)) \text{and} (overlap($image_i$,previous $discontinuity_{j-1}$) $< \epsilon$))
      }{
        
        
        
        \eIf{$simplifyGeometry$}{
           $newFace_i$ = convex hull of $discontinuity_j$
          
           depth($newFace_j$) = $classification_{i}[depth]$
         }{
           \For{each $vertex_{k}$ in convex hull of planar $discontinuity_j$ in .obj'
        }
        {
           depth($vertex_{k}$) = $classification_{i}[depth]$
        }
        }
        
      }
}

\hspace{1cm}

\textbf{Assumptions}: Scan of initial 3D reconstruction (.obj) with partials of object perimeter. Discontinuity detection using color and vertices on planes based on input. Future work, define a mapping from classifications to 3D geometry using RGB-D data,  tracking, or Iterative Closest Point (ICP)~\cite{10.1145/2047196.2047270}.

\textbf{Description}:
compare discontinuities in reconstructed geometry with EchoCNN inferences. If EchoCNN classifies a hole as closed, select planar vertices surrounding gap at EchoCNN estimated depth, create a new mesh based on its convex hull, and assign material based on EchoCNN estimated material.

\caption{Echoreconstruction via EchoCNN inference}
\label{alg:one}
\end{algorithm}

\begin{table*}
\centering
  \begin{tabular}{l|c|cc|cccc|ccc}
  \multicolumn{11}{c}{\bf Accuracy of Reflecting Sounds used for Open/Closed, Depth Estimation, and Material Classification in Real World Scenes} \\
  \hline
    & & \multicolumn{2}{c}{Open/Closed} & \multicolumn{4}{c}{Depth Estimation} & \multicolumn{3}{c}{Sound Material} \\ 
    Method        & Input               & Shower & Window & Overall & 3 ft & 2 ft & 1 ft & 
    Overall & Glass & Mirror \\
    \hline
    kNN~\cite{Cover:67} & A & 56.5\% & \textbf{100\%} & 21.3\% & 16\% & 21\% & 25\% & 44.0\% & 47.5\% & 52.4\%  \\ 
    Linear SVM~\cite{bottou-2010} & A & 61.5\% & 91.7\% & 37.6\% & 38\% & 32\% & 41\% & 51.9\% & 46.0\% & 57.1\% \\ 
    SoundNet5~\cite{aytar2016soundnet} & A & 45.2\% & 46.6\% & 39.7\% & 40\% & 71\% & 8\% & 71.0\% & 98.4\% & 1.6\% \\
    SoundNet8~\cite{aytar2016soundnet} & A & 50.7\% & 46.6\% & 42.5\% & 92\% & 0\% & 33\% & 44.4\% & 16.4\% & 85.7\% \\
    \textbf{EchoCNN-A} (Ours) & A & \textbf{71.2\%} & \textbf{100\%} & \textbf{71.8\%} & 86\% & 54\% & \textbf{76\%} & \textbf{77.4\%} & 62.3\% & \textbf{92.0\%} \\ 
    \hline
    ImageNet~\cite{NIPS2012_4824} & V & 78.1\% & 96.1\% & 45.2\% & 52\% & 83\% & 0\% & 80.6\% & 60.7\% & 100\% \\
    \hline
    Acoustic Classification~\cite{Schissler2018AcousticCA} & AV & N/A & N/A & N/A & N/A & N/A & N/A & \multicolumn{3}{c}{----------  48\% * ----------} \\
    \textbf{EchoCNN-AV Cat} (Ours) & AV & \textbf{100\%} & \textbf{100\%} & \textbf{89.5\%} & \textbf{95\%} & \textbf{100\%} & 73\% & \textbf{100\%} & \textbf{100\%} & \textbf{100\%} \\
    \textbf{EchoCNN-AV MFB} (Ours) & AV & \textbf{100\%} & \textbf{100\%} & 84.9\% & 54\% & \textbf{100\%} & \textbf{100\%} & 80.6\% & 60.7\% & \textbf{100\%} \\ 
    \hline
  \end{tabular}
    \caption{Multiple models (\textbf{ours is EchoCNN}) and baselines were evaluated for audio and audio-visual based scene reconstruction analysis. Overall, {\bf 71.2\%} of held out reflecting sounds and {\bf 100\%} of audio-visual frames were correctly classified as an open or closed interior surface (i.e. glass shower door). Open/closed classification is even higher for external facing windows due to outside noise. According to~\cite{Long2014}, 10 dB of exterior to interior noise reduction can be attributed to closed compared to open windows. Example shower and window views in \autoref{fig:allResults}. {\bf 71.8\%} of 1-second audio frames were correctly classified as 1 ft, 2 ft, or 3 ft away from surface based on audio alone; {\bf 89.5\%} when concatenating with its corresponding image. Finally, {\bf 77.4\%} and {\bf 100\%} of audio and audio-visual inputs correctly labeled the surface material. * According to~\cite{Schissler2018AcousticCA}, 48\% of the triangles in the scene are correctly classified, where its classification is more granular and covers more material classes.
  }
  \vspace*{-2em}
  \label{tab:model-eval}
\end{table*}

\vspace*{-1em}
\subsection{Depth filtering and planar inpainting}
The outputs of our EchoCNN inform enhancements for 3D reconstruction (\autoref{alg:one}). If depth jumps in the reconstruction are first classified as an open surface, then no change is required other than filtering loose geometry and small components. Otherwise, there is a planar discontinuity (e.g. window or mirror) that needs to be filled. With depth estimated by EchoCNN, we filter the initial 3D mesh to within a threshold of that depth. This gives us the plane size needed to fill. Finally, EchoCNN classifies its surface material.

\subsection{Implementation details}
\label{ImplementationDetails}
We implemented all EchoCNN and baseline models with Tensorflow~\cite{tensorflow2015-whitepaper} and Keras~\cite{chollet2015keras}. Training was performed using a TITAN X GPU running on Ubuntu 16.04.5 LTS. We used categorical cross entropy loss with Stochastic Gradient Descent optimized by ADAM~\cite{kingma2014method}. Using a batch size of 32, remaining hyperparameters were tuned manually based on a separate validation set. We make our real-world and synthetic datasets available to aid future research in this area.

\section{Datasets and Applications}
\label{Datasets}
Our audio-based EchoCNN-A and audio-visual EchoCNN-AV convolutional neural networks are trained across nine octave bands with center frequencies 63 Hz, 125 Hz, 250 Hz, 500 Hz, 1 kHz, 2 kHz, 4 kHz, 8 kHz, and 16 kHz. Training is done using these pulsed pure tone impulses along with experimenter hand clap. The hold out test data is comprised of sound sources excluded from training - white noise, experimenter whistle, and voice. The test set contains sound sources not in the training set to evaluate generalization.

\begin{figure}
  \includegraphics[width=0.45\textwidth]{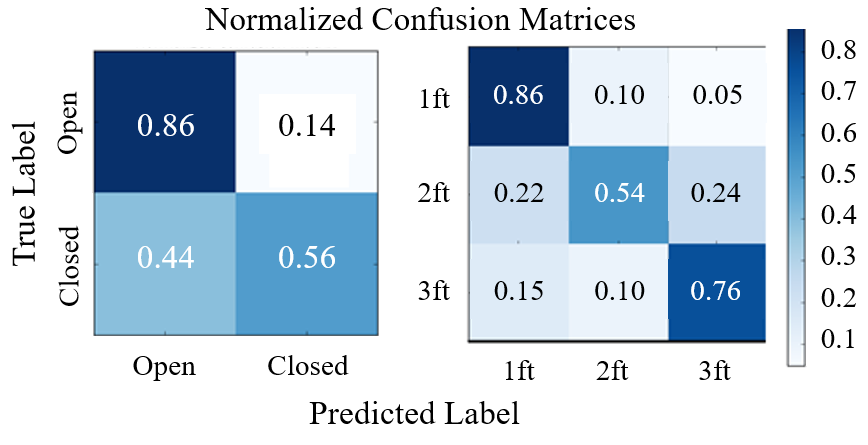}
  \vspace*{-1.5em}
  \caption{EchoCNN-A (\textit{Left}) Confusion matrix to classify open/closed for an interior glass shower door. Open predictions (86\%) were more accurate than closed (56\%). (\textit{Right}) Confusion matrix to classify depth from same interior glass door. Notice that our EchoCNN is learning to differentiate distance based on reflecting sounds from pulsed ambient waves of a smartphone.
  }
  \label{fig:confusionMatrix}
  \vspace*{-1.5em}
\end{figure}

\subsection{Real and synthetic datasets}
\label{RealAndSyntheticDatasets}
\textbf{Real}: training data is comprised of 1 second pulsed spectrograms (\autoref{fig:comparisonSpectrograms}) from recorded pure tones, experimenter hand claps, brownian noise, and pink noise (N=857). Training and test examples were collected via video recordings and labeled for material, open/closed, and in 1 ft depth increments based on the distance from the surface. Nine octaves of pure tones, hand claps, and white noise cover a disperse range of frequencies and were used to train our models. 

\begin{table*}
\centering
  \begin{tabular}{l|c|cc|cc|c}
  \multicolumn{7}{c}{\bf Accuracy of Reflecting Sounds used for Classification in Controlled Experiment (Gl = Glass)} \\
  \hline
    & & \multicolumn{2}{c}{Open/Closed} & \multicolumn{2}{c}{Depth Est. (+/- 0.5 m)} & \multicolumn{1}{c}{.} \\ 
    Method & Input & Thick Gl & Thin Gl & Thick Gl & Thin Gl &
    Material Est \\
    \hline
    kNN~\cite{Cover:67} & A & 53.8\% & 64.1\% & 11.5\% & 21.4\% & 66.5\%  \\ 
    Linear SVM~\cite{bottou-2010} & A & 54.7\% & 63.2\% & 11.5\% & 20.5\% & 61.1\% \\ 
    SoundNet5~\cite{aytar2016soundnet} & A & 60.0\% & 40.1\% & 18.8\% & 19.1\% & 67.4\% \\
    SoundNet8~\cite{aytar2016soundnet} & A  & 60.0\% & 42.6\% & 25.0\% & 19.1\% & 34.0\% \\
    \textbf{EchoCNN-A} (Ours) & A &  \textbf{61.1\%} & \textbf{65.1\%} & \textbf{44.4\%} & \textbf{44.6\%} & \textbf{68.1\%} \\ 
    \hline
    ImageNet~\cite{NIPS2012_4824} & V & 95.8\% & 80.8\% & 83.3\% & 66.7\% & 87.5\% \\
    \hline
    Acoustic Classification~\cite{Schissler2018AcousticCA} & AV & N/A & N/A & N/A & N/A & \multicolumn{1}{c}{ -- 48\% * -- } \\
    \textbf{EchoCNN-AV Cat} (Ours) & AV & 98.9\% & \textbf{100\%} & 99.4\% & 92.2\% & 76.6\% \\
    \textbf{EchoCNN-AV MFB} (Ours) & AV & \textbf{100\%} & \textbf{100\%} & \textbf{100\%} & \textbf{99.0\%} & \textbf{100\%} \\ 
    \hline
  \end{tabular}
    \caption{Multiple models (\textbf{ours is EchoCNN}) and baselines were evaluated for audio and audio-visual based scene reconstruction analysis. * According to~\cite{Schissler2018AcousticCA}, 48\% of the triangles in the scene are correctly classified, where its classification is more granular and covers more material classes. 
  Compared to other existing methods, ours is able to correctly classify open/closed surfaces and depth estimation at nearly 100\%, while achieving much higher accuracy in material estimation as well.}
  \vspace*{-2em}
  \label{tab:vir-model-eval}
\end{table*}

The hold out test dataset consists of 1-second pulsed spectrograms from recorded experimenter voice, whistle, chirp, and white noise (N=431). Voice and whistle recordings were chosen for the hold out test set to ease future transition to live and hands-free emitted sounds during reconstruction. Hold out test data is excluded from training and only evaluated during testing. While the same hold out sets were used for visual and audio-visual evaluation, unheard is not the same as unseen. Unheard audio can have the same visual appearance between training and test. Other new training and test datasets for visual and audio-visual methods will be future work.

\textbf{Synthetic}: 
The automated synthetic data collection was performed in Unreal Engine 4.25 where SteamAudio employs a ray-based geometric sound propagation approach, with support for dynamic geometry using the Intel Embree CPU-based ray-tracer. We describe similar prior work~\cite{GSound} here for more details on this approach. Given scene materials (e.g. carpet, glass, painted, tile, etc.), a sound source (e.g. voice), environmental geometry, and listener position, we generate impulse responses for a given scene of varying sizes. From each listener, specular and diffuse rays are randomly generated and traced into the scene. The energy-time curve for simulated impulse response $S_f(t)$ is the sum of these rays:
\vspace*{-0.5em}
\begin{equation}
    S_f(t) = \sum \delta(t-t_j)I_{j,f}
\end{equation}

\noindent where $I_{j,f}$ is the sound intensity for path j and frequency band f, $t_j$ is the propagation delay time for path j, and $\delta(t - t_j)$ is the Dirac delta function or impulse function. As these sound rays collide in the scene, their paths change based on absorption and scattering coefficients of the colliding objects. Common acoustic material properties can be referenced in~\cite{Egan1988}. We assume a sound absorption coefficient, $\alpha = 1.0$ for open windows.

Along with sound intensity $S_f(t)$, a weight matrix $W_f$ is computed corresponding to materials within the scene. Each entry $w_{f,m}$ is the average number of reflections from material m for all paths that arrived at the listener. It is defined as:

\begin{equation}
    w_{f,m} = \frac{\sum I_{j,f} d_{j,m}}{\sum I_{j,f}}
\end{equation}

\noindent where $d_{j,m}$ is the number of times rays on path j collide with material m, weighted according to the sound intensity $I_{j,f}$ of the path j. To mirror our real-world data, sound source directivity was disabled. Future work is needed to compare ambient and directed sound sources. This data may also be used for material sound separation.

Given a 720p 30fps video walkthrough of the virtual environment (VE) with the camera moving along a keyframed spline, we reconstruct the virtual scene by extracting the individual frames of the video and using Agisoft Metashape (v1.7)'s reconstruction pipeline to solve for each image's camera transform. Metashape, previously known as PhotoScan, is considered state-of-the-art in commercial photogrammetry software. The general process \cite{ linder2009digital} is: create a sparse point cloud containing only keypoints and solve for the transforms of cameras that can see the keypoints, create a dense cloud by matching more features between keypoint-seeing cameras and the rest of them, project the dense cloud depth data to each camera to build per-camera depth maps, use the depth maps to build a mesh, and create a texture map by projecting the image frames that best see each polygon onto the mesh \cite{catmull1974subdivision}. 

We disable motion blur of the camera in order to have more usable frames, but real camera data generally requires blurry frames to be removed to avoid noisy reconstructions, especially at low framerates. For accurate visual feedback of the specular surfaces, we also enable UE4's DirectX12 ray-tracing for reflective and translucent surfaces. 
We used a PC with the following specs for reconstruction: GTX 1080 GPU, i9-9900k CPU, 64gb RAM, Windows 10 x64. It takes approximately 2 hours to process a 720p image sequence of 2000 frames from start to finish with this setup.
We use three sections of the "HQ Residential House" environment on the Unreal Marketplace for synthetic data.
The kitchen and bathroom sequences result in about 2,000 images when they are extracted from the video at a step size of 3, and the master bedroom has about 4,000.

\begin{figure}
\vspace*{-0.5em}
  \includegraphics[width=0.38\textwidth]{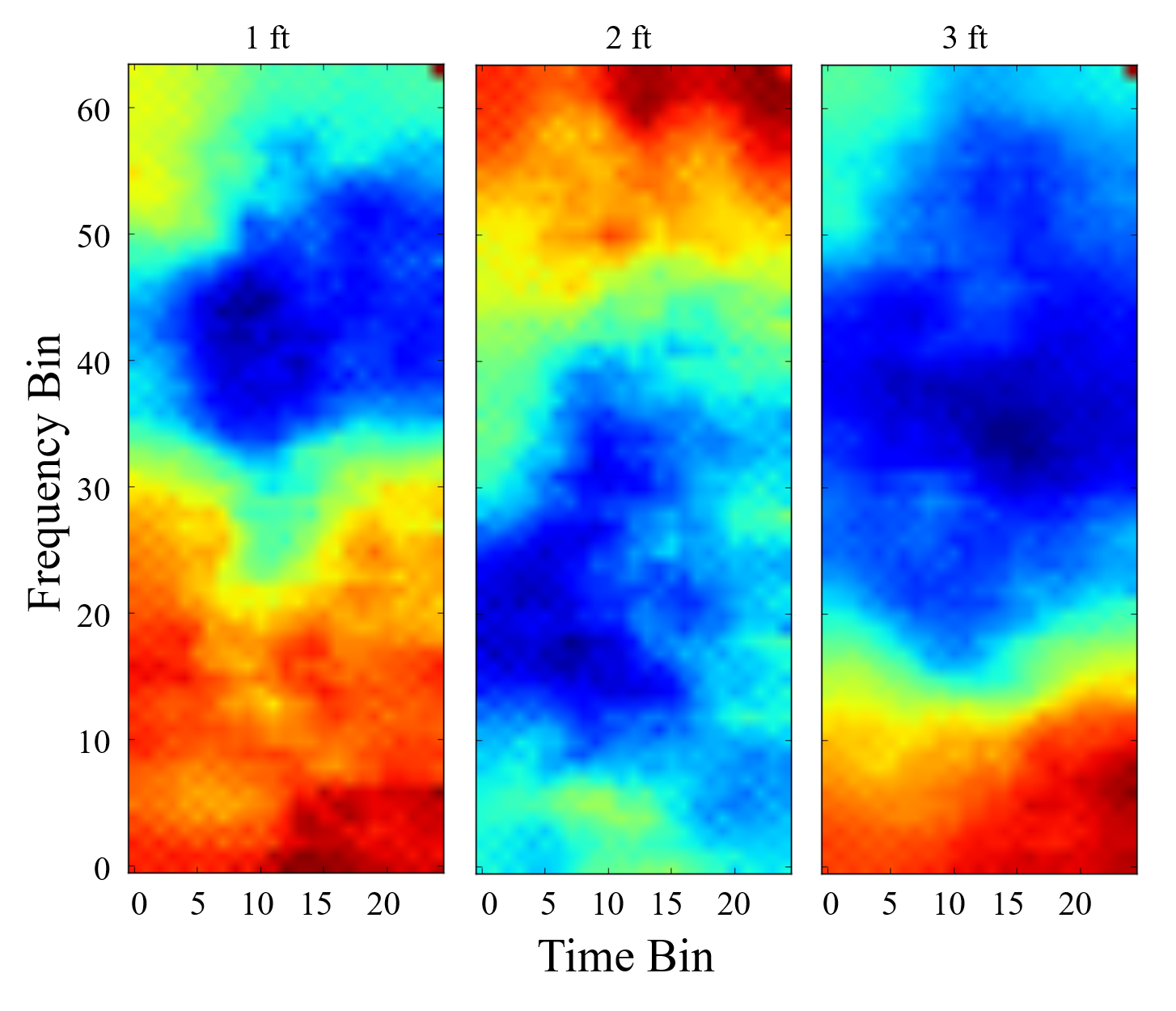}
  \vspace*{-1.5em}
  \caption{\textit{From left to right}: audio input (i.e. mel-scaled spectrogram) which would produce the highest activation for a given depth class from 1 ft, 2 ft, and 3 ft away from an object. Longer reverberation times tend to occur at lower frequencies (3 ft) than at high frequencies (1 and 2 ft) due to typical high frequency damping and absorption.}
  \label{fig:activationMaximization}
  \vspace*{-1.5em}
\end{figure}

\begin{figure*}
  \includegraphics[width=1\textwidth]{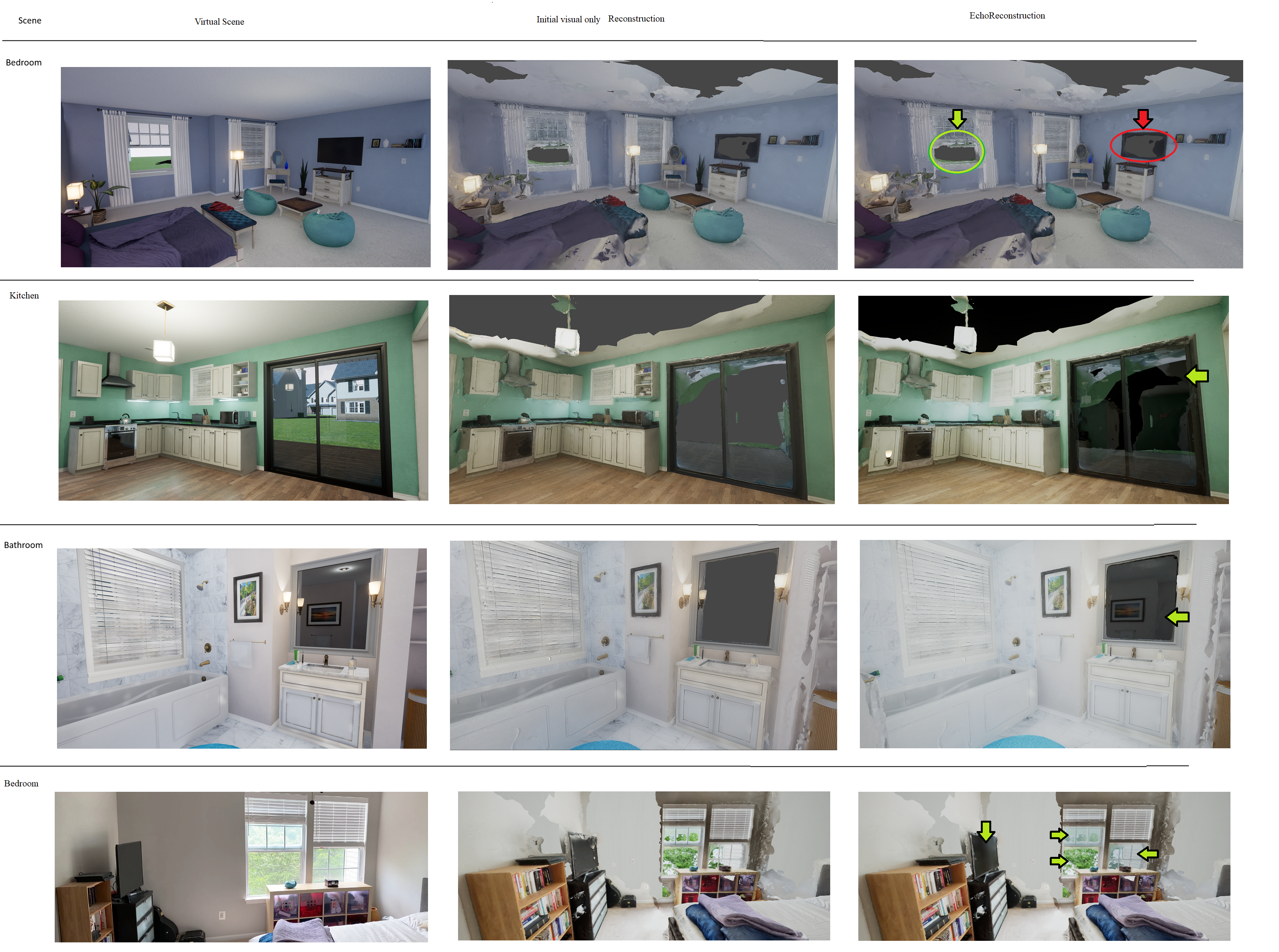}
  \vspace*{-2.5em}
  \caption{We evaluated our method using a controlled experiment of virtual scenes (\textit{row 1}: bedroom, \textit{row 2}: kitchen, and \textit{row 3}: bathroom) and applied to real world scenes (\textit{row 4}: physical bedroom).~\textit{Column 1}: input video recording reflecting sounds emitted from the camera as the scene is scanned.~\textit{Column 2}: an initial reconstruction using state-of-the-art commercial MetaShape application. 
  We tested glass, mirror, and other objects and surfaces within each scene at different depths, materials, and open/closed.~\textit{Column 3}: audio-augmented echoreconstruction after depth inpainting and semantic rendering is applied during post-processing. Failure case: while virtual bedroom window remained open (success, green), TV was not enhanced (failure, red) and in supplemental.}
  \label{fig:allResults}
  \vspace*{-0.5em}
\end{figure*}

\begin{figure}[h]
\vspace*{-1.5em}
  \includegraphics[width=0.45\textwidth]{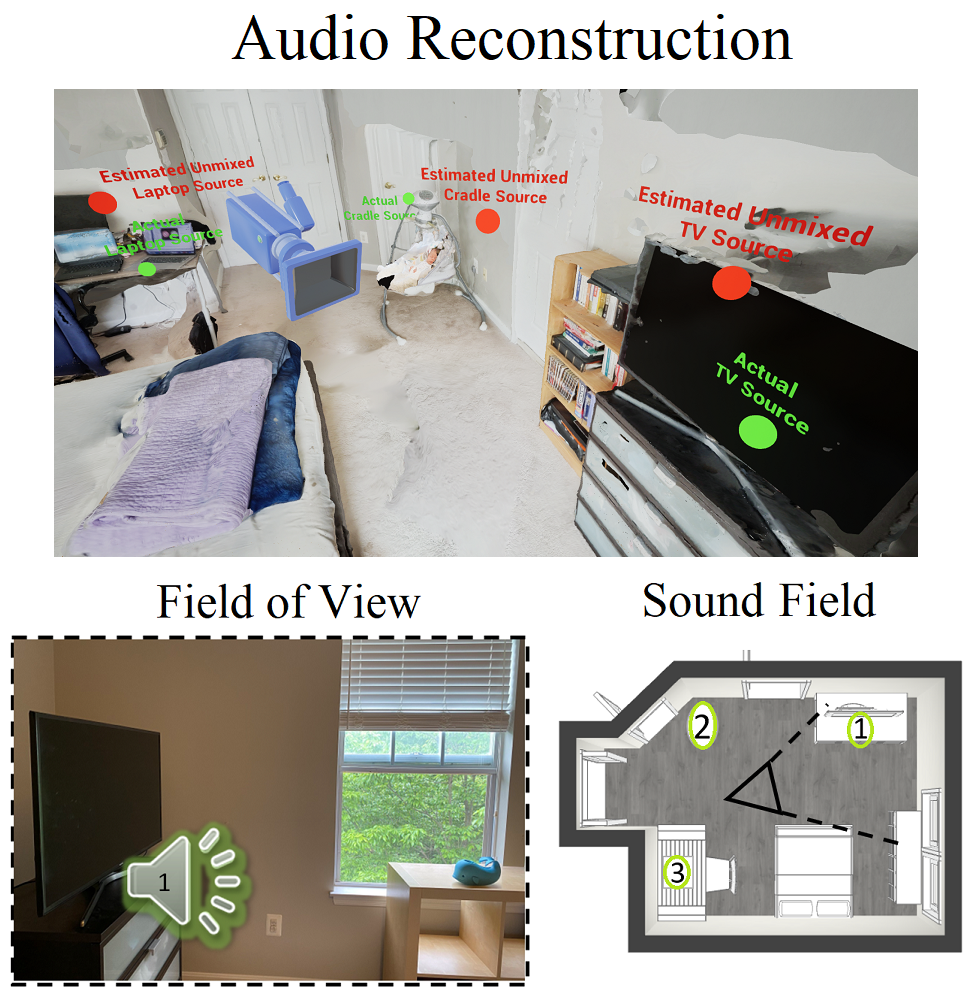}
  \vspace*{-1em}
  \caption{EchoCNN may also be used to reconstruct the audio of a virtual scene from a video of a physical room. Instead of depth estimation, our method can be trained to approximate sound source position, which is especially useful for objects that are outside of the camera field of view. Ground truth (green dots) and estimated (red dots) sound source placements are shown (top). Seen and heard sound source (TV) from the video capture is placed more accurately than unseen but heard sound sources (cradle and laptop). 
  Please see our supplementary video for a VR demo. 
  }
  \label{fig:audioReconstruction}
  \vspace*{-1em}
\end{figure}

\vspace*{-0.5em}
\subsection{Applications in VR Systems}
\label{Applications}
When using a head mounted display (HMD) users are alerted when approaching the boundaries in physical space. However, if room setup does not accurately reflect these boundaries or changes occur after setup, a user risks walking into unseen real-world objects such as glass and walls. Using our method, transmitted sound from the HMD could be used to locate physical objects and appropriately notify the user as an added safety measure. Audio directly from the real-world environment could also be used for depth estimation. The sounds unmixed and placed in the virtual environment, reconstructing both the scene geometry and sound sources (\autoref{fig:audioReconstruction}). Finally, seasonal variations in the 3D sound and visual reconstruction of a window open in the spring and closed in the winter also enhance the AR/VR experience.  See \autoref{fig:pointCloud} and supplementary demo video. 

\section{Experiments and Results}
\label{Results}
Overall, 71.2\% of hold out reflecting sounds and 100\% of audio-visual frames were correctly classified as an open or closed boundary in the home (\autoref{tab:model-eval}). 71.8\% of 1 second audio frames were correctly classified as 1 ft, 2 ft, or 3 ft away from the surface based on audio alone; 89.5\% when concatenating with its corresponding image. Finally, 77.4\% of audio and 100\% of audio-visual inputs correctly labeled the surface material.

ImageNet, a visual only baseline, is higher at 78.1\% than audio-only EchoCNN-A for open/closed classification. This is partly due to the fact that the hold out set was to test audio generalization (i.e. unheard sound sources). But unheard sound sources does not guarantee unseen visual data. Images similar to those found in training are present in test. A hold out set based on image (e.g. different depths) should be evaluated as future work.

\subsection{Experimental setup}
\label{ExperimentalSetup}
Listener (top smartphone, e.g. Galaxy Note 4) and sound source (bottom smartphone, e.g. iPhone 6) are separated vertically by 7 cm. Pulsed sounds are emitted 3 feet, 2 feet, and 1 feet away from the reconstructing surface. Three feet was selected to remain in the free field. Beyond that, there will be less noise reduction due to reflecting sounds in the reverberant field~\cite{Egan1988}. Within a few feet of the reconstructing surface also create finer detail reconstructions.

We labeled our data based on scene, sound source, and surface properties - type of surface, material, and depth from sound source. The training set included pulsed sounds of pure tone frequencies, a single hand clap, brownian noise, and pink noise. The hold out test set consisted of voice, whistle, chirp, and white noise. For rooms with different sound-absorbing treatments, our real-world recordings include a bedroom (e.g. carpet and painted) and bathroom (e.g. tiled).

\subsection{Activation Maximization}
\label{ActivationMaximization}
The objective of activation maximization is to generate an input that maximizes layer activations for a given class. This provides insights into the types of patterns the neural network is learning.~\autoref{fig:activationMaximization} shows the different inputs that would maximize EchoCNN activations for depth estimation. Notice lower frequencies tend to occur at 3 ft (longer reverberation times) than at 1 and 2 ft (high frequencies) due to typical high frequency damping and absorption.

\subsection{Analysis}
\label{Analysis}
Using audio, we noticed noise reduction between winter and spring due to more foliage on the trees. We also observed flutter echoes, which can be heard as a "rattle" or "clicking" from a hand clap and have been simulated in spatial audio~\cite{Halmrast2019AVS}. They became more pronounced the closer to the wall surface in the bathroom scene. Background UV textures are placed at a fixed 1 ft (0.3 m) behind estimated surface depth. Audio unable to augment failure cases of the shower from initial RGB-based reconstructions using either~\cite{Tanskanen2013LiveM3} or~\cite{Metashape20}. We leave calculating the background depth as future work. We compare our 3D reconstructions to depth estimates based on related work.

\subsection{Results by source frequency and object size}
\label{ResultsBySourceFrequencyAndObjectSize}

We evaluate a range of source frequencies to account for different sound wave behavior based on the size of the reconstructing objects. For example, if an object is much smaller than the wavelength, the sound flows around it rather than scattering~\cite{Long2014}:
\vspace*{-0.75em}
\begin{equation}
    \lambda = \frac{c}{f}
\vspace*{-0.5em}
\end{equation}
\noindent where $\lambda$ is wavelength (ft) of sound in air at a specific frequency, $f$ is frequency (1 Hz), and $c$ is speed of sound in air (ft/s). Dynamically setting source frequency based on object size would be future work.

\vspace*{-0.5em}
\section{Conclusion and Future Work}

To the best of our knowledge, our work introduces the first audio and audio-visual techniques for enhancing scene reconstructions that contain windows and mirrors. Our smartphone prototype and staged EchoReconstruction pipeline emits and receives pulsed audio from a variety of sound sources for surface detection, depth estimation, and material classification. These classifications enhance scene and object 3D reconstruction by resolving planar discontinuities caused by open spaces and reflective surfaces using depth filtering and planar filling. Our system performs well compared to baseline methods given experiment results for real-world and virtual scenes containing windows, mirrors, and open surfaces. We intend to publicly release our real and synthetic audio-visual ground truth data in addition to reflection separation data (direct, early, or late reverberations) for future research.

This work offer many exciting possibilities in teleimmersion, teleconferencing, and many other AR/VR applications, where the improved quality and accessibility of scanning a room with mobile phones can significantly enhance the presence of users. This work can be integrated into VR headsets with cameras and microphones, enabling a remote walk-through of a space like museums, architectures, future homes, or a cultural/heritage site, for example. The scanning process could give feedback in real time so the user wearing the HMD could see the quality of the reconstruction and move to areas with issues, like mirrors and open windows. 

\noindent
{\bf Future Work}: To further extend this research, performing audio emission, reception, and 3D reconstruction {\em simultaneously} in real time instead of having a staged approach is an alternative to explore. This approach could enable mapping classifications to 3D geometry more densely than fusing RGB-D, tracking, or Iterative Closest Point (ICP)~\cite{10.1145/2047196.2047270}. An integrated approach, such as a multi-scale neural network~\cite{DBLP:journals/corr/EigenF14} using audio for coarse and visual for finer predictions, may not only be more efficient but also more effective by using audio feedback as part of the reconstruction code.  Another possible avenue of exploration is to investigate the impact of live audio for training and/or testing our neural network variations. With a defined set of output classes for EchoCNN, alternative baselines such as Non-Negative Matrix Factorization (NMF), source separation techniques, and the pYIN algorithm~\cite{Mauch2014} to extract the fundamental frequency $f_0$, i.e. the frequency of the lowest partial of the sound, are suggested as future directions. Finally, our current implementation holds out voice and whistle data, which is different from the audio used during training. However, unheard sounds does not equate to unseen images. Therefore, some insights can be possibly gained by experimenting with a different training dataset for testing audio-only, visual-only, and audio-visual methods.

%
%
%
%

\bibliographystyle{ACM-Reference-Format}
\bibliography{samplebody-journals}

\appendix
\end{document}